%% file: template.tex
\documentclass[a4paper]{article}

\usepackage{INTERSPEECH2022}

\usepackage{multirow}
\usepackage{booktabs}
\usepackage[urlcolor=blue]{hyperref}
\pdfoutput=1

\title{Enhancing Word-Level Semantic Representation via Dependency Structure for Expressive Text-to-Speech Synthesis}

\name{Yixuan Zhou$^{1,\ddagger}$\thanks{$^{\ddagger}$Work conducted when the first author was intern at Tencent.}, Changhe Song$^{1,\dagger}$ \thanks{$\dagger$ Equal contribution.}, Jingbei Li$^1$, Zhiyong Wu$^{1,2,*}$ \thanks{* Corresponding author.},\\  Yanyao Bian$^{3}$, Dan Su$^{3}$, Helen Meng$^{2}$}
\address{
    $^1$ Shenzhen International Graduate School, Tsinghua University, Shenzhen, China\\
    $^2$ The Chinese University of Hong Kong, Hong Kong SAR, China\\
    $^3$ Tencent AI Lab, Tencent, Shenzhen, China}
\email{zhouyx20@mails.tsinghua.edu.cn, zywu@sz.tsinghua.edu.cn
}

\begin{document}
\maketitle

\input{abstract}
\input{introduction}

\input{methodology}

\input{experiments}

\input{conclusions}

%\appendix
\newpage

\bibliographystyle{IEEEtran}
\bibliography{references}

\end{document}

%% file: abstract.tex
\begin{abstract}
Exploiting rich linguistic information in raw text is crucial for expressive text-to-speech (TTS).
As large scale pre-trained text representation develops, bidirectional encoder representations from Transformers (BERT) has been proven to embody semantic information and employed to TTS recently.
However, original or simply fine-tuned BERT embeddings still cannot provide sufficient semantic knowledge 
that expressive TTS models should take into account.
In this paper, we propose a word-level semantic representation enhancing method based on dependency structure and pre-trained BERT embedding. 
The BERT embedding of each word is reprocessed considering its specific dependencies and related words in the sentence, to generate more effective semantic representation for TTS.
To better utilize the dependency structure, relational gated graph network (RGGN) is introduced to make semantic information flow and aggregate through the dependency structure.
The experimental results show that the proposed method can further improve the naturalness and expressiveness of synthesized speeches on both Mandarin and English datasets\footnote{\href{https://thuhcsi.github.io/interspeech2022-dependency-semantic-tts}{https://thuhcsi.github.io/interspeech2022-dependency-semantic-tts}}.

\end{abstract}

%% file: introduction.tex
\section{Introduction}

\begin{figure*}[!htb]
	\centering
	\includegraphics[width=0.93\linewidth]{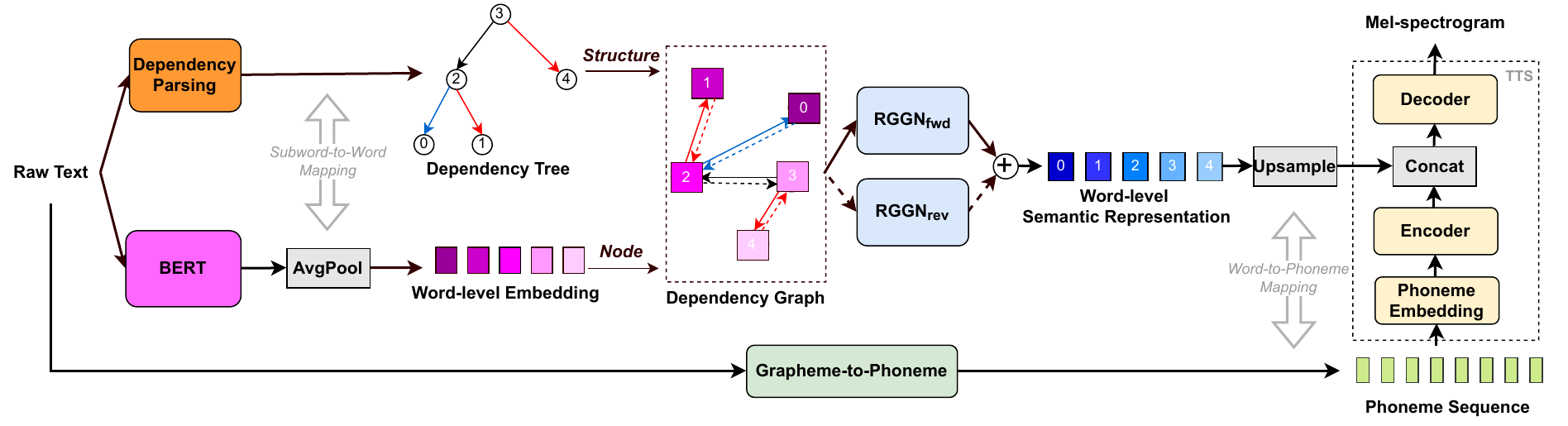}
	\caption{Main framework of the proposed method.
	In the middle dashed box, the dependency graph of a sentence consists of two relational graphs: forward relational graph (represented by solid arrows) and reverse relational graph (represented by dashed arrows).
	$RGGN_{fwd}$ and $RGGN_{rev}$ represent the relational gated graph networks dealing with forward and reverse graphs respectively.}
	\label{fig:model_structure}
    \vspace{-0.5em}
\end{figure*}

Text-to-speech (TTS), as an integral component of human-computer interaction frameworks, aims to generate natural speech with rich expressiveness from given text \cite{taylor2009text}.
As deep learning techniques develop,
neural TTS systems \cite{wang2017tacotron, shen2018natural, ren2020fastspeech} have shown great advantages over the conventional TTS systems \cite{zen2009statistical, takamichi2015naist}.
Despite nowadays neural TTS systems can synthesize speech with good intelligibility and naturalness, there is still much room for improvement in the expressiveness of generated speech.

There are two major ways to enhance the expressiveness of speech: i) modeling from reference speech; ii) predicting from input text.
The first way usually leverages a reference encoder to extract style representations from speech, such as global style token (GST) \cite{wang2018style} or variational autoencoder (VAE) \cite{zhang2019learning}.
These methods can generate more expressive speech, but require a reference utterance or manually-determined token weights, which suffers from training/inference mismatch since ground-truth reference speech used in training are usually unavailable in inference.
To avoid this problem, the second way investigates to model and predict the variation of generated speech from text directly.
In this regard, how to exploit the rich linguistic information\footnote{\href{https://en.wikipedia.org/wiki/Linguistics}{https://en.wikipedia.org/wiki/Linguistics}} in raw text (including syntax, semantics, pragmatics, etc.) corresponding to the variation of speech is crucial to handling the one-to-many mapping problem of expressive TTS \cite{tan2021survey}.

Some studies devote efforts to improving the expressiveness of TTS system by introducing syntactic information.
To provide better syntactic representations as additional input for TTS, 
the word relation based features (WRF) \cite{guo2019exploiting} and the syntactic parse tree traversal \cite{song2021syntactic} are successively proposed.
Besides, GraphTTS \cite{sun2020graphtts} and GraphSpeech \cite{liu2020graphspeech} incorporate syntactic information through graph networks, both of which optimize the structure of TTS models by explicitly connecting input phonetic embeddings with syntactic relations.
All these approaches have demonstrated that introducing syntactic structure of text to TTS, by either providing extra input or using dedicated modeling methods, can improve the prosody of generated speech, especially in terms of pauses and rhythms.

Whereas, it is difficult to fully characterize the rich variations in human speech by just using syntactic information without considering the semantics of a sentence.
As large-scale pre-trained text representation develops, bidirectional encoder representations from Transformers (BERT) \cite{devlin2018bert} has been proven to embody semantic information and employed in many downstream tasks.
Inspired by this,
BERT is firstly incorporated to TTS as an additional input and shows gains in mean opinion scores of the synthesized speech \cite{hayashi2019pre, kenter2020improving}.
Similar method further verifies the ability of BERT to improve prosody on the Chinese multi-speaker TTS task \cite{xiao2020improving}.
In addition, there are some attempts that use BERT for predicting style variations in speech \cite{zhang2021extracting}.
The success of these methods indicates that exploiting the latent semantic information from BERT does show potential benefits for expressive speech synthesis.

However, original or simply fine-tuned BERT embeddings still cannot provide sufficient semantic knowledge that expressive TTS models should take into account.
Such knowledge includes but is not limited to the descriptions of inter-word relations that can help clarify the meaning of the sentence.
BERT is trained on large-scale data in a self-supervised manner, where only raw plain texts are involved.
Although the attention heads of BERT could capture several specific relation types somewhat well \cite{htut2019attention}, they do not reflect the full extent of the significant  semantic structure knowledge.
In recent decades, dependency parsing (DP) is developed for semantic analysis and semantic structure extraction, where words are directly connected by dependency links with labels \cite{kubler2009dependency, zhang2020survey}.
These dependency relations between words are associated with semantic information, and GraphSpeech has demonstrated that solely using these dependencies can help improve the prosody.

To further improve the expressiveness of synthesized speech, we propose a word-level semantic representation enhancing method based on dependency structure and pre-trained BERT embedding. 
The BERT embedding of each word is reprocessed considering its specific dependencies and related words in the sentence, to generate more effective semantic representations for TTS.
To better utilize the dependency structure, the relational gated graph network (RGGN) is introduced to make semantic information flow and aggregate through the dependency structure. 
The experimental results on both Mandarin and English datasets show that the proposed method outperforms the baseline using vanilla BERT embedding and another proposed BLSTM-based enhancing method in terms of naturalness and expressiveness.
And the case study indicates that semantic representations generated by the proposed method are more diverse and distinguished, which benefits expressive text-to-speech synthesis.

%% file: methodology.tex
\section{Methodology}

The main framework of the proposed method is illustrated in Fig.\ref{fig:model_structure}.
After constructing the dependency graph with semantic information from the raw text, two individual RGGNs are used to generate enhanced word-level semantic representations that are passed as additional input to TTS for boosting the expressiveness of synthesized speech.

\begin{figure}[t]
  \centering
  \includegraphics[width=0.9\linewidth]{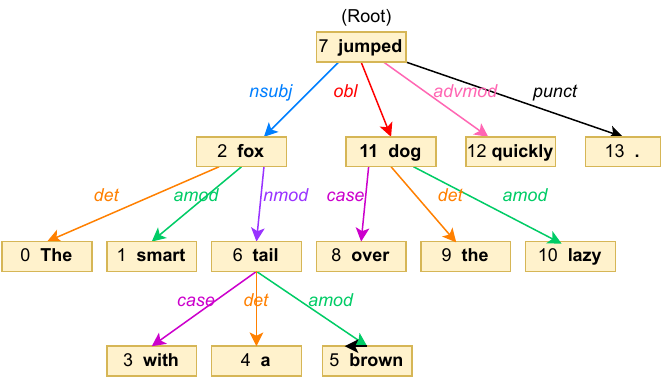}
  \caption{Dependency tree of ``The smart fox with a brown tail jumped over the lazy dog quickly." 
  Different colored edges represent different kinds of dependency relations.
  For example, the edge label between ``jumped" and ``fox" is ``nsubj", pointing from ``jumped" to ``fox", which means ``fox" is the nominal subject of the dependency head ``jumped".
  }
  \label{fig:dependencyparsing_exp}
  \vspace{-0.5em}
\end{figure}

\subsection{Dependency graph with semantic information}

In a dependency tree, bilexicalized dependencies are used to describe the relations between words for semantic analysis, as shown in Fig.\ref{fig:dependencyparsing_exp}.
We use a directed graph to represent a sentence in the form of the dependency tree, defined as $\mathcal{G}=(\mathcal{V}, \mathcal{E})$.
Node $v\in \mathcal{V}$ represents the token with semantic information, usually a word.
Directed edge is defined as the pair $e=(v_i, v_j)\in \mathcal{E}$, representing a particular dependency relation from node $v_i$ (head) to node $v_j$ (dependent).

The dependency graph is constructed from raw text by pre-trained dependency parsing model and BERT model.
Because the nodes of the dependency tree are at the word level while the tokens of BERT are at the subword level, the sentence $T$ must be represented in these two levels respectively:
\begin{equation}
T =\left\{
\begin{aligned}
& [w_1, w_2, ..., w_n] = \mathcal{W} \\
& [sw_1, sw_2, ..., sw_m] = \mathcal{SW} 
\end{aligned}
\right.
\end{equation}
where $w_i$ is the $i$-th word, $sw_j$ is the $j$-th subword, $n$ and $m$ are the number of words and subwords in the sentence.
Then we can obtain the dependency structure and the BERT embeddings:
\begin{equation}
\begin{split}
    [\mathbf{b}_{1}, \mathbf{b}_{2}, ..., \mathbf{b}_{m}] &= BERT(\mathcal{SW}) \\
    \mathcal{E}_{dep} &= DP(\mathcal{W}) 
\end{split}
\end{equation}
where $\mathbf{b}_{j}$ is the $j$-th subword-level embedding of BERT outputs, and $\mathcal{E}_{dep}$ is the set of directed edges with labels
from dependency parsing that can be regarded as the dependency structure of $\mathcal{W}$.
The outputs of BERT model are then converted into word-level embeddings by the average pooling with a subword-to-word mapping, that can be formulated as:
\begin{equation}
\begin{split}
    \mathbf{v}_{i} = Avg&Pool(\mathbf{b}_{i_1}, \mathbf{b}_{i_2}, ..., \mathbf{b}_{i_k})  \\
   \mathcal{V}_{sem} &= [\mathbf{v}_{1}, \mathbf{v}_{2}, ..., \mathbf{v}_{n}] 
\end{split}
\end{equation}
where ${i_1}$ to ${i_k}$ are the indices of subwords belonging to the $i$-th word.
$\mathbf{v}_i$ is the word-level embedding of the $i$-th word, which composes the node set $\mathcal{V}_{sem}$ with semantic information.

Considering the interaction between dependents and heads, 
we introduce another directed edge set $\mathcal{E}_{rev}$ by reversing the direction of each edge in $\mathcal{E}_{dep}$ and retaining the edge labels.
Meanwhile, we rename the original $\mathcal{E}_{dep}$ as $\mathcal{E}_{fwd}$.
$\mathcal{E}_{fwd}$ and $\mathcal{E}_{rev}$ are treated as two individual sets of relational types, and used to construct the forward and reverse relational graphs:
\begin{equation}
\begin{split}
    \mathcal{G}_{fwd}=&(\mathcal{V}_{sem}, \mathcal{E}_{fwd}) \\
    \mathcal{G}_{rev}=&(\mathcal{V}_{sem}, \mathcal{E}_{rev})
\end{split}
\end{equation}
$\mathcal{G}_{fwd}$ and $\mathcal{G}_{rev}$ together compose
the dependency graph of a sentence.
The semantic information of words will flow to the related nodes directly or indirectly through the structure of $\mathcal{G}_{fwd}$ or $\mathcal{G}_{rev}$, which will be described in detail in the next section.

\subsection{Relational gated graph network}
The relational gated graph network (RGGN) aims to enhance semantic representations based on two relational graphs in the dependency graph.
Considering the advantages of gated graph neural network (GGNN) in long-term propagation of information flow \cite{li2015gated}, we extend it to RGGN inspired by \cite{schlichtkrull2018modeling}.

For each relational graph, the hidden state $h_i^0$ is first initiated by the word-level embedding $\mathbf{v}_i\in \mathcal{V}_{sem}$ for node $i$.
Then the propagation step computes node representation for each node through several iterations, formulated as:
\begin{equation}
\begin{split}
    & a_i^t = \sum_{j \in N(i)} W_{eij}h_j^t \\
    & h_i^{t+1} = GRU(a_i^t, h_i^t)
\end{split}
\end{equation}
where $N(i)$ is the set of neighbors of node $i$, $W_{eij}$ denotes the weight parameters of edge from node $i$ to $j$ in propagation determined by its label,
$h_i^{t}$ and $h_i^{t+1}$ denote hidden states of node $i$ in the $t$-th and the $(t$+1$)$-th iteration, 
$a_i^t$ is the weighted sum of the hidden states $h_j^t$ of the neighbors of node $i$,
and $GRU(\cdot)$ denotes the gated aggregator which updates hidden state $h_i^t$ incorporating information both from the neighbors and the previous iteration step of node $i$.
The hidden states of the nodes after the last iteration step are then passed to a fully-connected layer.

The RGGNs dealing with the forward and reverse relational graphs are denoted as $RGGN_{fwd}$ and $RGGN_{rev}$, respectively.
They have the same structure but independent non-shared parameters.
The final semantic representations are obtained as:
\begin{equation}
\begin{split}
    \{r_{fwd}^i\}_{i=1, 2, ..., n} &= RGGN_{fwd}(\mathcal{G}_{fwd}) \\
    \{r_{rev}^i\}_{i=1, 2, ..., n} &= RGGN_{rev}(\mathcal{G}_{rev}) \\
\end{split}
\end{equation}
\begin{equation}
    \begin{split}
        r_{bi}^i & = r_{fwd}^i \oplus r_{rev}^i, i\in \{1, 2, ..., n\} \\
    R_{bi} &= [r_{bi}^1, r_{bi}^2, ..., r_{bi}^n]
    \end{split}
\end{equation}
where $r_{fwd}^i$ and $r_{rev}^i$ are the representations of the $i$-th word in $\mathcal{W}$ learned from the forward and reverse relational graphs respectively, $\oplus$ is the element add operation of $r_{fwd}^i$ and $r_{rev}^i$ to get the final semantic representation $r_{bi}^i$,
and $R_{bi}$ is the word-level semantic representation sequence obtained by reordering $r_{bi}^i$, that is consistent with $\mathcal{W}$.

\subsection{TTS with semantic representations}

For TTS model, Tacotron 2 \cite{shen2018natural} is adopted as the synthesizer in our work.
We use phoneme sequences as TTS input, which is obtained from the raw text by grapheme-to-phoneme (G2P) conversion.
Word-level semantic representations $R_{bi}$ are duplicated to match the number of phonemes each word contains, and then concatenated with the TTS encoder outputs.
Finally, the concatenation outputs are passed to the attention-based decoder of Tacotron 2 to generate mel-spectrograms.

%% file: experiments.tex
\section{Experiments}

\subsection{Experimental setup}

For comparison, four systems are implemented with PyTorch backend\footnote{\href{https://github.com/thuhcsi/tacotron}{Implemented based on: https://github.com/thuhcsi/tacotron}}.  
\textbf{(1) Vanilla}: The vanilla Tacotron 2 with phoneme-input only.
\textbf{(2) BERT}: a method of integrating BERT into Tacotron 2 is proposed in \cite{hayashi2019pre}. 
To be consistent with other systems,
we adapt the method by replacing subword-level BERT embedding with word-level through the average pooling mentioned in 2.1 and using phonemes rather than characters as input of Tacotron 2.
\textbf{(3) BERT-Dep(RGGN)}: Our proposed method that use both $RGGN_{fwd}$ and $RGGN_{rev}$ to enhance semantic representations.
\textbf{(4) BERT-Dep(BLSTM)}: To verify the effectiveness of RGGN, we propose another semantic representation enhancing method with bidirectional long short-time memory (BLSTM) \cite{hochreiter1997long}. 
The directly connected dependency label (through a lookup table) and the dependency head position (through a position encoding) of each word are regarded as dependency features and concatenated with word-level BERT embedding, and then passed to a BLSTM to obtain the enhanced semantic representations.

For the details in RGGN, the dimension of the GRU layer and the FC layer are both set to 768, consistent with the output dimension of BERT.
There are 55 categories of edge labels according to the types of dependencies.
The number of iterations is set to 5 which is enough to reach the furthest related word in a common sentence.
RGGN is implemented based on deep graph library (DGL) \cite{wang2019deep}.
In the training stage, RGGN and Tacotron 2 are jointly trained without any special tricks.

To present the generality of the proposed method, we train and evaluate on both Mandarin and English single-speaker datasets.
For Mandarin, we use a 12-hour Mandarin female speaker corpus from DataBaker \cite{databaker}.
For English, we use a 24-hour English female speaker corpus LJSpeech \cite{ljspeech}.
After 100 sentences are reserved as the test set, 95\% of the sentences are used for training and the rest are used for validation for two datasets.
Following GraphSpeech, we use Stanza \cite{qi2020stanza} as the pre-trained dependency parsing model.
And we use the pre-trained parameters of language-dependent BERT-base models \cite{berten, bertzh}.
The learning rate is set to $10^{-3}$.
We train all the models for 140K steps with a batch size of 32 on an NVIDIA P40 GPU.
Besides, we use a well-trained HiFi-GAN vocoder \cite{kong2020hifi} to synthesize waveforms from mel-spectrograms.

\subsection{Subjective evaluations}
We employ mean opinion score (MOS) evaluations of speech expressiveness and naturalness by subjective listening tests.
20 sentences are selected randomly from the test set and Internet, which are varied in length and content.
20 listeners without hearing impairs participated in the test, and rated from 1 to 5.

\begin{table}[th]
\renewcommand{\arraystretch}{1}
  \caption{The MOS results with 95\% confidence intervals on LJSpeech (EN) and DataBaker (CN) datasets.}
  \label{tab:mos}
  \centering
  \begin{tabular}{ l@{}l  r r}
    \toprule
    \multicolumn{2}{c}{\textbf{Model}} & \multicolumn{1}{c}{\textbf{MOS (EN)}} & \multicolumn{1}{c}{\textbf{MOS (CN)}}\\
    \midrule
    Vanilla & & $3.240\pm0.111$ & $3.215\pm0.101$ ~~~               \\
    BERT    & & $3.469\pm0.105$ & $3.488\pm0.100$ ~~~               \\
    BERT-Dep(BLSTM)     & & $3.531\pm0.117$ & $3.519\pm0.103$ ~~~              \\
    BERT-Dep(RGGN)      & & $\mathbf{3.903\pm0.103}$ & $\mathbf{3.927\pm0.088}$ ~~~              \\
    \bottomrule
  \end{tabular}
\end{table}

As shown in Table \ref{tab:mos}, the proposed BERT-Dep(RGGN) receives the highest MOS of about 3.91, exceeding Vanilla by 0.69, BERT by 0.44, and BERT-Dep(BLSTM) by 0.39, respectively. 
It demonstrates that our proposed method significantly improves the naturalness and expressiveness of generated speeches both for Mandarin and English datasets. 
Compared with Vanilla, three systems that used BERT perform better, indicating that introducing semantic information is effective. 
BERT-Dep(RGGN) achieves great improvement in MOS while BERT-Dep(BLSTM) shows few gains over BERT, 
which demonstrates the effectiveness of introducing RGGN to utilize dependency structure, 
while simply using dependency information as an additional feature may not work.

\subsection{Ablation study}

Three ablation studies are conducted by removing $RGGN_{fwd}$, $RGGN_{rev}$, and edge labels, respectively in the proposed method.
And we employ comparison mean opinion score (CMOS) evaluations to evaluate the expressiveness of generated speeches. 
As shown in Table \ref{tab:cmos}, these removals of the specific design all lead to performance degradation, where removing $RGGN_{fwd}$ results in the worst performance (even generates some bad cases in LJSpeech) and removing $RGGN_{rev}$ results in a little or no drop, indicating that the forward relational graph contributes more to semantic representation enhancing for expressive TTS.
For reference, we also conduct a CMOS evaluation between the proposed method and BERT-Dep(BLSTM) at the same time. 
It can be seen that removing edge labels still outperforms BERT-Dep(BLSTM), which reflects the advantages of introducing the structure itself from dependency parsing.

\begin{table}[th]
\renewcommand{\arraystretch}{1}
  \caption{The CMOS results on LJSpeech (EN) and DataBaker (CN) datasets.}
  \label{tab:cmos}
  \centering
  \begin{tabular}{ l@{}l  r r}
    \toprule
    \multicolumn{2}{c}{\textbf{Model}} & \multicolumn{1}{c}{\textbf{CMOS (EN)}} & \multicolumn{1}{c}{\textbf{CMOS (CN)}}\\
    \midrule
    BERT-Dep(RGGN) & & $0$ & $0$ ~~~     \\
    \midrule
    \hspace{3mm}- $RGGN_{fwd}$     & & $-1.057$ & $-0.528$ ~~~ \\
    \hspace{3mm}- $RGGN_{rev}$     & & $-0.267$ & $-0.097$ ~~~ \\
    \hspace{3mm}- Edge labels     & & $-0.181$ & $-0.313$ ~~~ \\
    \midrule
    BERT-Dep(BLSTM)    & & $-0.429$ & $-0.626$ ~~~ \\
    \bottomrule
  \end{tabular}
  \vspace{-1.0em}
\end{table}

\subsection{Case Study}
\begin{figure}[htb]
    \centering
    \includegraphics[width=0.8\linewidth]{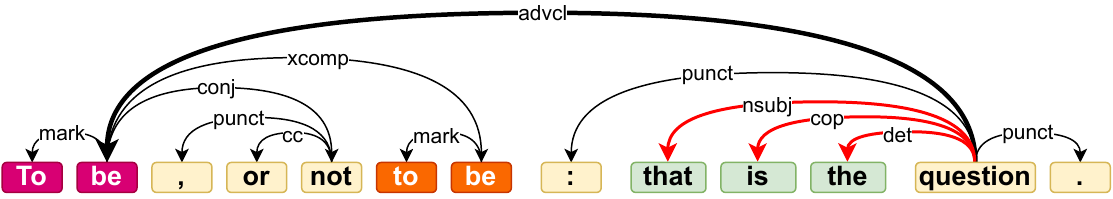}\\
    (a) dependency tree of the sentence
    \includegraphics[width=0.6\linewidth,
    height=0.22\linewidth]{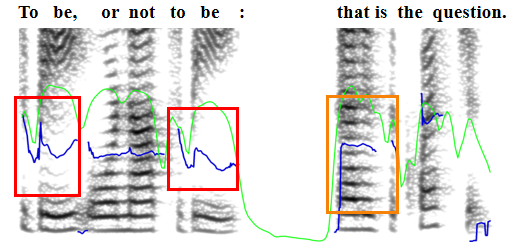}\\
    (b) baseline BERT\\
    \includegraphics[width=0.6\linewidth, height=0.22\linewidth]{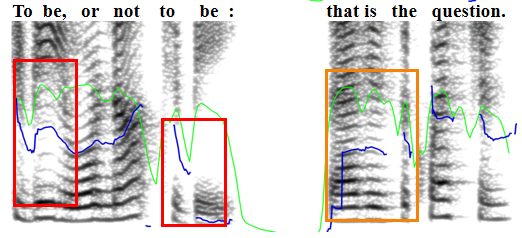}\\
    (c) proposed BERT-Dep(BLSTM)\\
    \includegraphics[width=0.6\linewidth, height=0.22\linewidth]{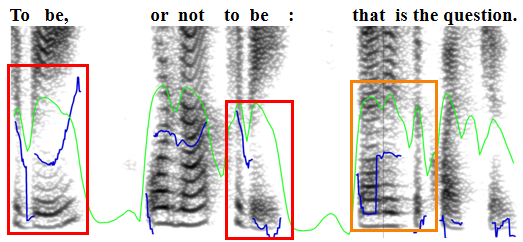}\\
    (d) proposed BERT-Dep(RGGN)
      \caption{Mel-spectrogram, pitch (blue curve) and intensity (green curve) of the speeches synthesized by three systems for the sentence ``To be, or not to be: that is the question."}
    \label{fig:casestudy}
    \vspace{-0.5em}
\end{figure}

We make a case study by comparing the mel-spectrogram, pitch contour and intensity of the speeches generated from different systems, as shown in Fig.\ref{fig:casestudy}. 
These features are extracted by Praat\footnote{\href{http://www.praat.org/}{http://www.praat.org/}}.
As can be seen, the pronunciations of two ``to be"s in the proposed BERT-Dep(RGGN) are more diverse compared with the other two systems.
The two ``to be"s of the baseline BERT are similar and relatively flat in terms of pitch.
The first ``to be" has a higher pitch than the second one in both two proposed methods, but their pitch trends are more distinguished in BERT-Dep(RGGN).
This difference is attributed to the different dependency structures of two ``be"s, as shown in Fig.\ref{fig:casestudy}(a).
The pause corresponding to the first comma is more obvious in BERT-Dep(RGGN).
For ``that" and ``question", it can be observed that ``question" is emphasized in BERT and BERT-Dep(BLSTM), which may due to its richer semantic information than ``that" from pre-trained BERT.
Instead, ``that" rather than ``question" is emphasized in BERT-Dep(RGGN) considering the semantic information of ``question" is passed to ``that is the" through the dependency structure, 
resulting in a more natural expression.

\begin{figure}[!htb]
	\centering
	\includegraphics[width=0.7\linewidth]{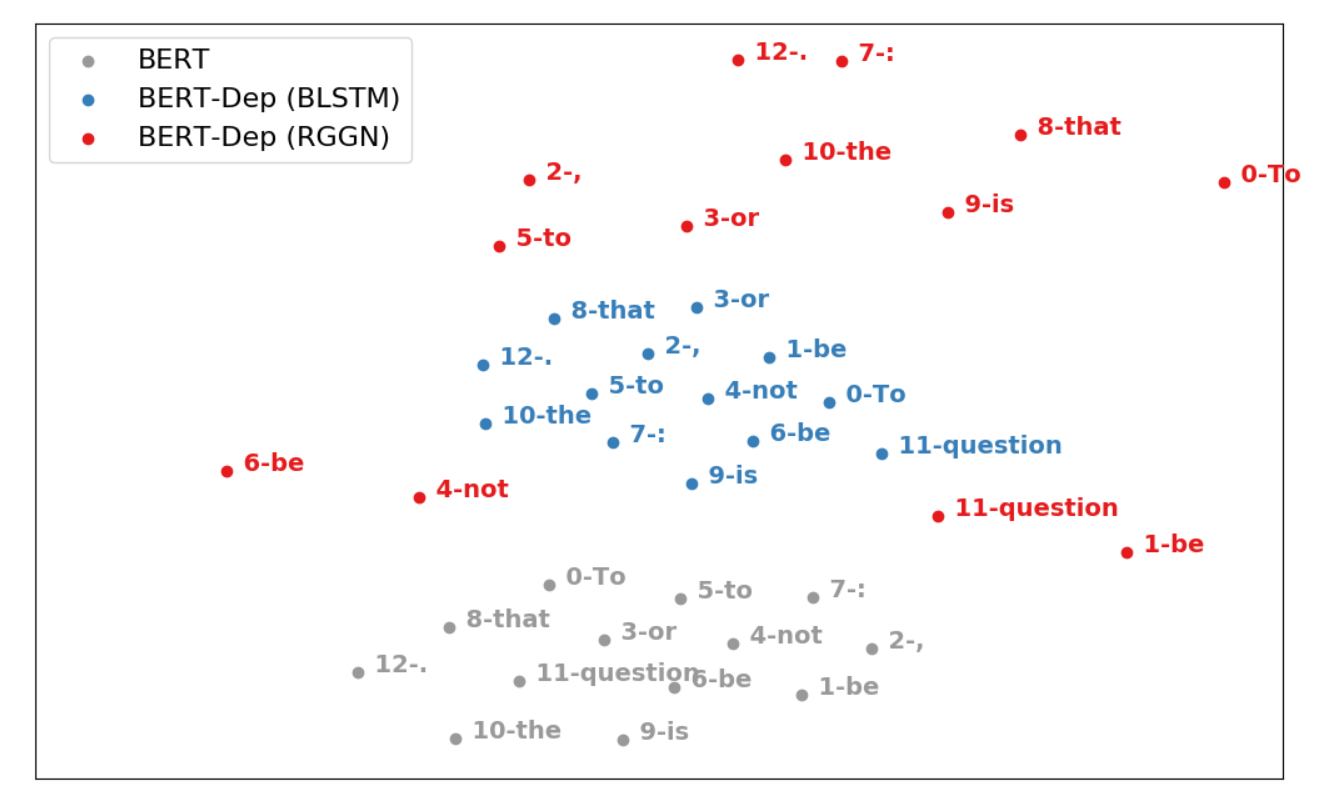}
	\caption{T-SNE visualization of word-level semantic representations learned by different systems. The number before ``-" is the index of the word or punctuation in the sentence. }
	\label{fig:tsne}
	\vspace{-1.0em}
\end{figure}

Furthermore, we visualize the semantic representations by 2D t-SNE \cite{van2008visualizing} as shown in Fig.\ref{fig:tsne}.
The semantic representations obtained by the proposed method are more scattered than the other two settings, which demonstrates the effectiveness of the dependency structure driven by RGGN in incorporating these relations between words to enhance the diversity of original BERT embeddings.
For example, the two ``be"s are further apart in BERT-Dep(RGGN), corresponding to the phenomena in the generated speech mentioned above.

%% file: conclusions.tex
\section{Conclusions}
In this study, we propose a word-level semantic representation enhancing method.
The BERT embedding of each word is reprocessed considering its specific dependencies and related words in the sentence, to generate more effective semantic representations for expressive TTS.
The experimental results show that the proposed method can further improve the naturalness and expressiveness of synthesized speeches on both Mandarin and English datasets.

\textbf{Acknowledgement}: This work is supported by National Natural Science Foundation of China (NSFC) (62076144), Shenzhen Science and Technology Innovation Committee (WDZC20200818121348001), Tencent AI Lab Rhino-Bird Focused Research Program (JR202143) and Tsinghua University - Tencent Joint Laboratory.